\pdfoutput=1
\documentclass[runningheads]{llncs}

\usepackage{latexsym}
\usepackage{float}
\usepackage{graphicx}
\usepackage{subcaption}
\usepackage[T1]{fontenc}
\usepackage{subcaption}
\usepackage{wrapfig}
\usepackage{amsmath}
\usepackage{multirow}
\usepackage{booktabs}
\usepackage{graphicx}
\usepackage{url}
\usepackage{xcolor}
\usepackage[utf8]{inputenc}
\usepackage{enumitem}
\usepackage{microtype}
\usepackage{makecell}
\usepackage{adjustbox}
\usepackage{changepage} 
\usepackage[most]{tcolorbox}
\usepackage{multicol}
\usepackage{xcolor}
\newcommand{\Ours}{\textsc{Eeg}-Defender}

\title{Defending against Jailbreak through Early Exit Generation of Large Language Models}

\author{Chongwen Zhao\orcidID{0009-0003-2434-2785} \and
Zhihao Dou \and
Kaizhu Huang\orcidID{0000-0002-3034-9639}}

\authorrunning{C. Zhao et al.}
\titlerunning{Defending against Jailbreak through Early Exit Generation}

\institute{Duke Kunshan University, Kunshan, Jiangsu, China.\\
\email{chongwen.zhao, zhihao.dou, kaizhu.huang@dukekunshan.edu.cn} }

\begin{document}
\maketitle
\begin{abstract}
% Large Language Models (LLMs) are increasingly attracting attention in various applications. However, some malicious users try to use LLMs for illegal activities, such as drug production and fake news making. People use "Alignment" technology to avoid this circumstance. Recent study shows that the alignment can be circumvent by well-designed prompts or adversarial suffix, which is called Jailbreak. 
Large Language Models (LLMs) are increasingly attracting attention in various applications. 
Nonetheless, there is a growing concern as some users attempt to exploit these models for malicious purposes, including the synthesis of controlled substances and the propagation of disinformation. In an effort to mitigate such risks, the concept of "Alignment" technology has been developed. However, recent studies indicate that this alignment can be undermined using sophisticated prompt engineering or adversarial suffixes, a technique known as "Jailbreak." 
Our research takes cues from the human-like generate process of LLMs. We identify that while jailbreaking prompts may yield output logits similar to benign prompts, their initial embeddings within the model's latent space tend to be more analogous to those of malicious prompts. Leveraging this finding, we propose utilizing the early transformer outputs of LLMs as a means to detect malicious inputs, and terminate the generation immediately. We introduce a simple yet significant defense approach called \Ours{} for LLMs. We conduct comprehensive experiments on ten jailbreak methods across three models. Our results demonstrate that \Ours{} is capable of reducing the Attack Success Rate (ASR) by a significant margin, roughly 85\% in comparison with 50\% for the present SOTAs, with minimal impact on the utility of LLMs.

% EEG-Defend is capable of diminishing the Attack Success Rate (ASR) by a significant margin, roughly 90%, without compromising the utility and effectiveness of the LLMs.

% In this paper, inspired by the nature of language, we discover that although the output logits of jailbreaking prompts are similar to those of benign prompts, their early-stage embeddings in the latent space are closer to harmful counterparts. This insight enables us to utilize the early transformer outputs of LLMs to identify malicious inputs and halt generation instantly. \\ 
% We introduce \EEG-Defender for LLMs in this article. We conduct exhaustive experiments on ten jailbreaking methods across three models. Our results demonstrate that \EEG-Defender can reduce the Attack Success Rate (ASR) by approximately 90\% while preserving the helpfulness of the LLM.


\textbf{\textcolor{red}{Warning: this paper may contain offensive prompts and outputs.}} 
\end{abstract}

% \vspace{5pt}
\section{Introduction}\label{sec:introduction}

% Large Language Models (LLMs) are increasingly getting attention and application. The most successful and prominent application is chatbots, such as ChatGPT~\cite{achiam2023gpt} and Llama~\cite{touvron2023llamaopenefficientfoundation}. They are usually trained on large datasets collected from the web, enabling them to generate human-like and high-quality responses to user prompts. However, these datasets often contain inappropriate and/or harmful   content~\cite{ouyang2022traininglanguagemodelsfollow}, including biased, illegal, pornographic, and fraudulent material~\cite{weidinger2021ethical}. To avoid generating such responses to users, researchers have developed several alignment algorithms, including RLHF~\cite{ouyang2022traininglanguagemodelsfollow}, SFT~\cite{wei2022finetunedlanguagemodelszeroshot}, and PRO~\cite{song2024preferencerankingoptimizationhuman}. With these alignment algorithms, chatbots can refuse to respond to naive harmful prompts.

Large Language Models (LLMs) are garnering unprecedented attention and application in the field of artificial intelligence, with chatbots such as ChatGPT \cite{achiam2023gpt} and Llama \cite{touvron2023llama} standing out as notable examples. However, an inherent challenge arises because these models could generate inappropriate and potentially harmful content, including biased, unlawful, pornographic, and fraudulent material \cite{weidinger2021ethical}. To mitigate such risks and to steer LLM-generated responses away from these issues, researchers have innovated a series of alignment algorithms \cite{ouyang2022traininglanguagemodelsfollow,wei2022finetunedlanguagemodelszeroshot,song2024preferencerankingoptimizationhuman}. Through the implementation of these algorithms, chatbots have been empowered to discern and tactfully refuse to generate outputs in response to prompts that naively seek to elicit potentially harmful content.

% These sophisticated systems are trained on extensive datasets from the web, which imbues them with an impressive ability to generate responses that closely mimic human-like conversation in both quality and essence. 

\begin{figure*}[t]
    \centering
    \begin{subfigure}[b]{0.4\textwidth}
        \centering
        \includegraphics[width=\textwidth]{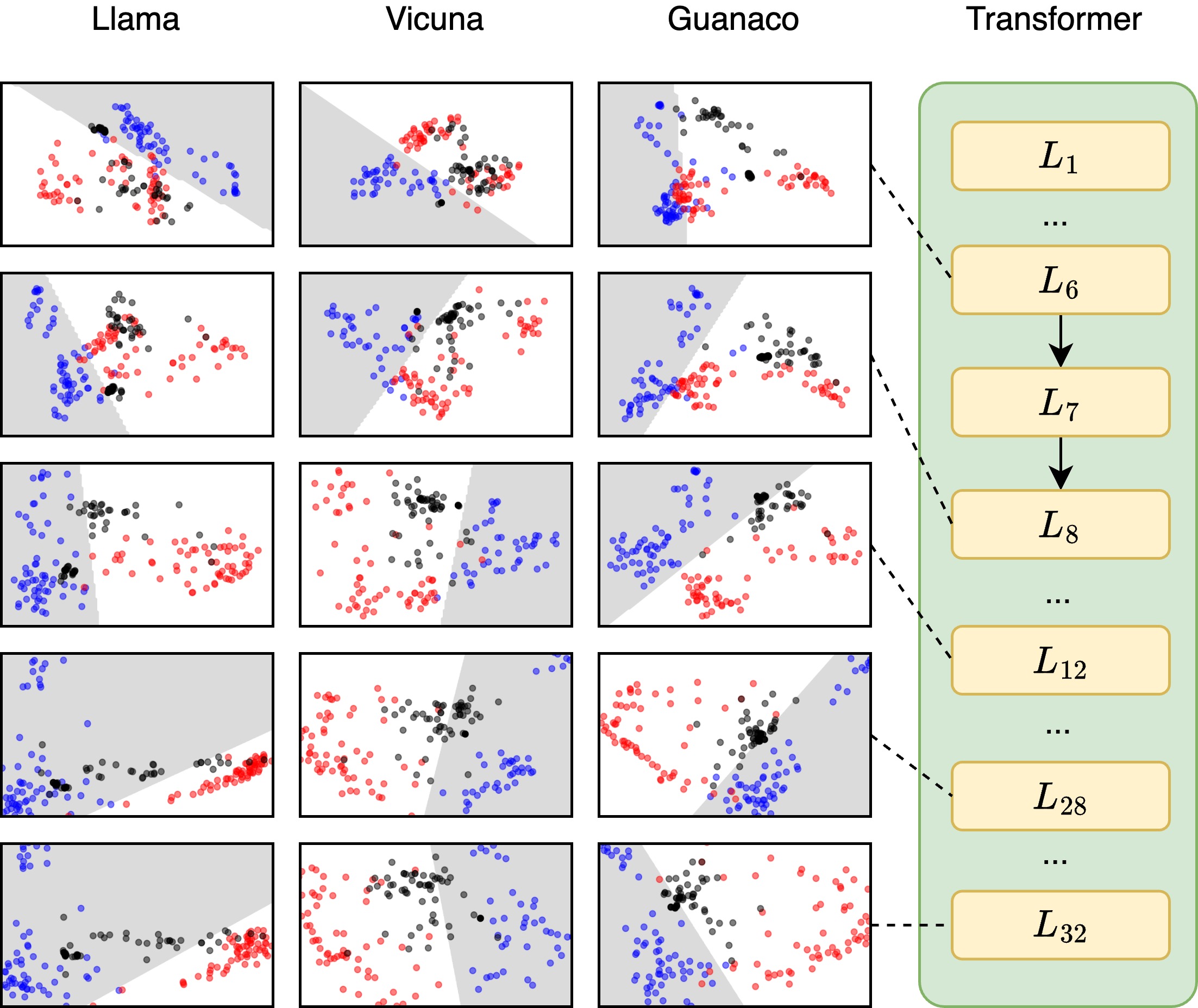}
        \caption{Model Embedding visualization.}
        \label{fig:pca}
    \end{subfigure}
    \hfill
    \begin{subfigure}[b]{0.55\textwidth}
        \centering
        \includegraphics[width=\textwidth]{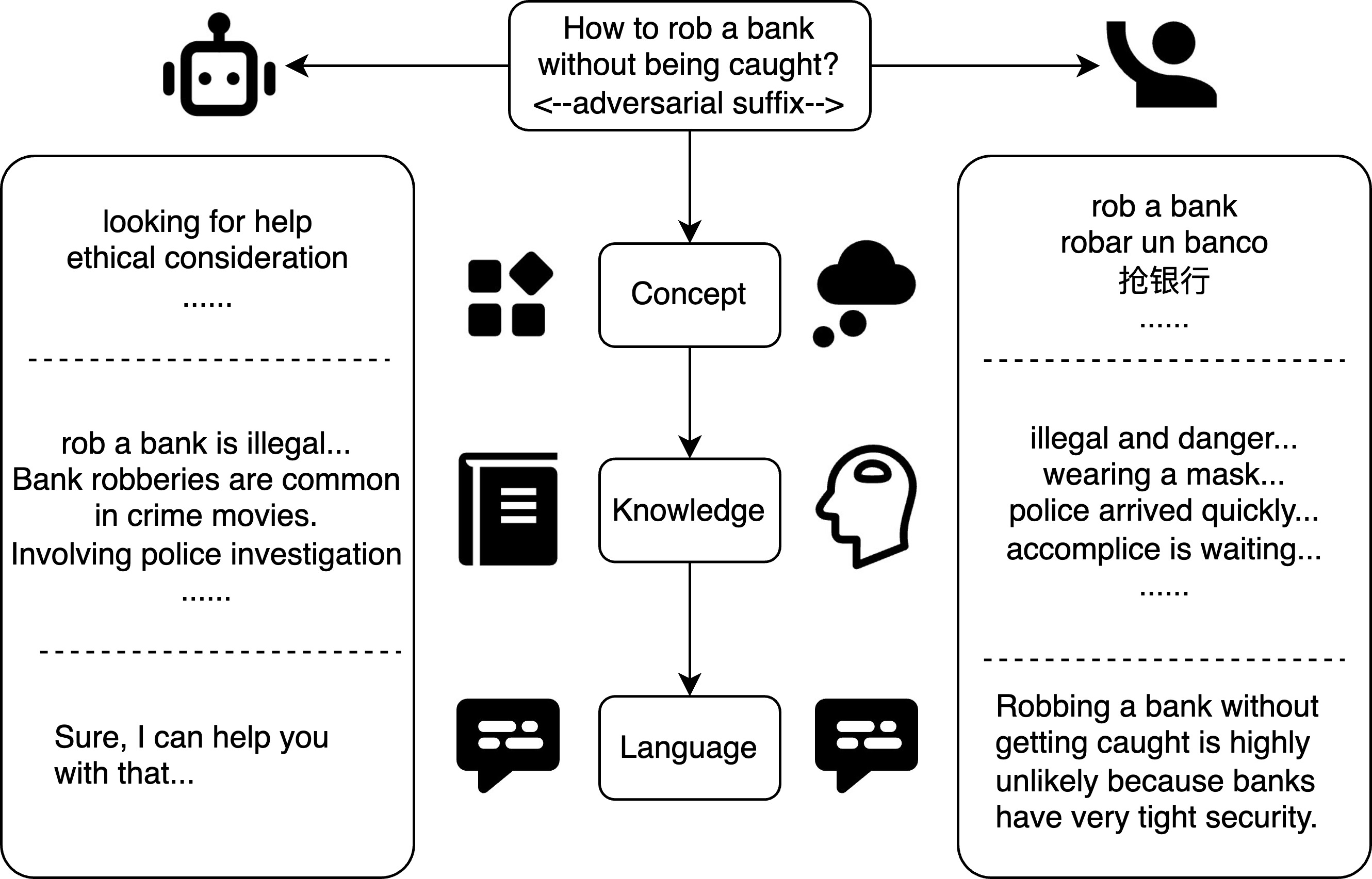}
        \caption{Language generation process.}
        \label{fig:con}
    \end{subfigure}
    \caption{Our insight stems from the human-like generation process of LLMs. Humans first develop an idea, then recall memories and organize language. Similarly, LLMs identify functions, retrieve knowledge in the middle layers, and generate language in the later layers. We found that in the early and middle layer latent space, \textbf{jailbreak prompts (black dots)} are more closer to \textbf{harmful prompts (red dots)} than to \textbf{benign prompts (blue dots)}.}
    \label{fig:explain_example}
\end{figure*}

% \begin{figure*}[t]
%     \centering
%     \includegraphics[width=0.85\textwidth]{Fig/conception.jpg}
%     \caption{}
%     \label{fig:explain_example}
%     \vspace{-10pt}
% \end{figure*}

% (1) 当前防守方法的问题：只考虑了最后一层，第一层是另外一个问题。
% (2) 我们认为只考虑最后一层是不合适的。我们假设前面层是更重要。
% (3) 我们用empirical visulization去验证前面层更重要。（4个图 Llama）
% (4) 这个idea事实上可以由人的认知来support.

More recently, it has been however discovered that well-designed jailbreak prompts can circumvent such alignment, posing new challenges for building stricter safety barriers \cite{GCG,liu2024autodangeneratingstealthyjailbreak,wei2023jailbrokendoesllmsafety}. Meanwhile, efforts to defend against jailbreaks are ongoing. Prompt-based methods \cite{zhang2024defending,self_reminders,jain2023baseline,wei2023jailbreak,llamaguard} approach defense by manipulating or detecting user prompts. However, these methods are impractical since they degrade significantly in utility \cite{xu2024safedecodingdefendingjailbreakattacks}. As a result, researchers turn to decoding-based defense methods \cite{robey2024smoothllmdefendinglargelanguage,cao2024defendingalignmentbreakingattacksrobustly,xu2024safedecodingdefendingjailbreakattacks,zhao2024defendinglargelanguagemodels}. Instead of directly accessing prompts, decoding-based defense methods leverage the model's internal properties. Since these methods can maintain high model functionality, decoding-based defense methods have shown promise in defending against jailbreak attacks.

% For instance, \cite{GCG} and \cite{liu2024autodangeneratingstealthyjailbreak} investigate adversarial prompts to maximize the likelihood of generating an affirmative response prefix. \cite{wei2023jailbrokendoesllmsafety} identify several mechanisms that can induce jailbreaks. These methods enable chatbots to generate harmful content.

Unfortunately, current decoding-based defense technologies are insufficient. Studies show that present defense methods could only reduce the Attack Success Rate (ASR) by around 50\% against jailbreak prompts \cite{xu2024comprehensivestudyjailbreakattack}. Approaches like RA-LLM \cite{cao2024defendingalignmentbreakingattacksrobustly} and Smooth-LLM \cite{robey2024smoothllmdefendinglargelanguage} propose generating responses multiple times with random dropouts to defend against character-sensitive adversarial suffix attacks. However, they are less effective against prompt crafting attacks, which typically involve character-insensitive prompts. SafeDecoding \cite{xu2024safedecodingdefendingjailbreakattacks} aims to increase the likelihood of disclaimer generation artificially, but in practice, it fails to effectively reduce ASR in models with stronger safety barriers.

In response to the drawbacks of existing decoding-based defense methods, we revisit the functions of different layers in LLMs. The method mentioned in \cite{todd2024function} reveals that the initial layers specialize in triggering specific tasks. The middle layers act as repositories of knowledge and shape the emotional tone of the output \cite{zhou2024alignmentjailbreakworkexplain,zhao2023explainabilitylargelanguagemodels}. Subsequent layers are where the refinement of the language output occurs \cite{fan2024layersllmsnecessaryinference}. Given that language only affects how we deliver, but not the semantics of expression \cite{Fedorenko2024}, we postulate that LLMs process all prompts similarly when recognizing functions in the initial layers and accessing stored knowledge in the middle layers.

To validate our postulation, we conduct a series of analyses. First, our results in Section \ref{classify_exp} demonstrate that the classifiers trained on the initial layers achieve over 80\% accuracy in detecting fail-to-refuse harmful prompts. More intuitively, as illustrated in Figure \ref{fig:pca}, our empirical visualization shows that starting from the early layers of models (e.g., layer 6 and layer 8), embeddings of jailbreak prompts aligned with harmful prompts. In the middle layers (e.g., layer 12), where LLMs retrieve information, jailbreak embeddings shift towards benign embeddings slightly, and by the later layers (e.g., layers 28 and 32), they become increasingly aligned with benign embeddings. Ultimately, the jailbreak embeddings are either distributed throughout the space (as seen with Llama2) or distributed with the decision boundary (as seen with Vicuna and Guanaco), complicating the model in recognizing jailbreak status.

Remarkably, the process by which large language models generate responses closely mirrors how humans organize language. To structure language output, humans first form an idea \cite{piaget}, then draw upon experiences and memories \cite{memoriesandlanguage,tulving1972episodic}. Finally, language serves as a conduit for conveying information \cite{thinkingandlanguage,Fedorenko2024}. As such, we argue that the focus may be placed on the early or intermediate layers rather than the latter or even final layers, which are overemphasized by current methods.

Based on this insight, we propose a simple yet novel framework for defending jailbreak, utilizing \textbf{E}arly \textbf{E}xit \textbf{G}eneration to defend against jailbreak, namely \textbf{\Ours{}}. Specifically, we exploit benign prompts and rejected harmful prompts as anchors for each layer's output. If the embeddings from the early and middle layers are sufficiently similar to the harmful anchor, the model will refuse the user’s request. We evaluate three popular LLMs: Llama2, Vicuna, and Guanaco. Despite its simplicity, our results show that \Ours{} significantly outperforms all five baselines under most conditions, achieving approximately an 85\% reduction rate in ASR while maintaining high functionality on benign prompts. Notably, \Ours{} requires no fine-tuning of the original LLM and incurs minimal additional computational cost compared to existing defense methods, making it seamlessly integrable into current workflows.

In summary, our contributions are three-fold:
\vspace{-2pt}
\begin{itemize}[leftmargin=*]
\item \textbf{Human-like generation process of LLMs.} Our study reveals that the generation process of LLMs parallels human language organization, a notable phenomenon not addressed in previous research.
\item \textbf{Latent space mechanism of jailbreak.} We empirically demonstrate that embeddings of jailbreak prompts in the early and middle layers closely resemble those of harmful prompts, but shift towards benign prompts in the later layers.
\item \textbf{Defend jailbreak through early exit.} We propose \Ours{}. \Ours{} reduces Attack Success Rate (ASR) by approximately 85\% against existing jailbreak methods, with near-zero computational cost.

\end{itemize}

% \vspace{5pt}
\section{Background and Related Work}

\subsection{Preliminaries}
We first define the key notations used in this paper.
\newline \textbf{Embeddings.} In LLMs, the embedding $e$ refers to the outputs produced by the transformer layers. Let $x_{1:s}$ denote a $s$-length user prompt, the LLM will generate output starting from $x_{s+1}$. In the final layer $n$, the embedding $e_{n}$ is used to generate the probability of the next token $x_{s+1}$ to $x_{1:s}$ by:
$$
p_\theta(x_{s+1} | x_{1:s}) = \text{softmax}(We_{n}),
$$
where $\theta$ denotes a language model and $W$ represents the $k\times m$ projector matrix that maps the embedding space $R^{m}$ to the token space $R^{k}$.
\newline \textbf{Jailbreak.} Jailbreak process aims to construct an adversarial prompt to elicit a harmful output of LLMs. Let $h$ denote a harmful question, and $\theta$ denote a language model. The process of jailbreak is to find $x_{1:s}$ by solving:
$$
    \max_{x_{1:s}} \quad  \prod_{i=0}^{|x_{s+1:}|} p_\theta\left(x_{s+i} \mid x_{1: s+i}\right),
$$
where $\exists i, j \text{ such that } x_{i:j} = h$ and $x_{s+1:}$ starting with "Sure, here is ..." instead of a disclaimer or rejection response. 
\newline \textbf{Harmful Prompts and Jailbreak Prompts.} Harmful prompts are straightforward requests for harmful or illegal behavior. In contrast, jailbreak prompts are complex which may include repressive denial and virtual context, or adversarial suffixes. Well-aligned LLMs can reject naive harmful prompts but may still accept jailbreak prompts. 
\newline \textbf{Benign Prompts.} These are user prompts that adhere to ethical guidelines, requesting assistance from LLMs without violating any norms.

\subsection{LLM Jailbreak}
Jailbreak attacks are generally categorized into prompt crafting and token optimizing. 
\newline \textbf{Prompt Crafting.} The study in \cite{wei2023jailbrokendoesllmsafety} found that LLMs are often vulnerable to jailbreaks due to competing objectives and mismatched generalizations. They proposed 30 jailbreak methods to elicit harmful responses from GPT and Claude. To alleviate the burden of manually designing jailbreak prompts, recent studies \cite{yu2024gptfuzzerredteaminglarge,mehrotra2024treeattacksjailbreakingblackbox,chao2024jailbreakingblackboxlarge} have introduced automated frameworks for generating effective attacks against LLMs. 
\newline \textbf{Token Optimizing.} In a white-box setting, attackers have access to the gradients of LLMs, allowing them to optimize prompts to increase the likelihood of generating affirmative responses. GCG attack \cite{GCG} achieved jailbreak by optimizing an adversarial suffix to minimize the loss of the desired prefix of outputting. The AutoDAN attack \cite{liu2024autodangeneratingstealthyjailbreak}. constructed prompts that can pass perplexity testing  Additionally, ICL attack \cite{qiang2024hijackinglargelanguagemodels} combined In-Context Learning with model gradients to distract the model’s attention and generate harmful content.

\subsection{Jailbreak Defense}
Defense strategies against jailbreaks can be broadly categorized into prompt-based methods and decoding-based methods.
\newline \textbf{Prompt-based Defense.} Detecting harmful content at the prompt level is an effective preventive strategy. To this end, Llama Guard \cite{llamaguard}, OpenAI \cite{openaimoderation}, and Perspective \cite{Perspective} provide APIs for content moderation. Beyond detection, prompt manipulation has also been explored: Zhang et al. \cite{zhang2024defending} appended prompts with safety-oriented instructions, while Self-Reminders \cite{self_reminders} leveraged psychological cues via system prompts to encourage responsible responses. Jain et al. \cite{jain2023baseline} proposed three strategies—perplexity detection, paraphrasing, and reorganization—but noted high false positive rates, limiting practical deployment.
\newline \textbf{Decoding-based Defense.} Some jailbreak prompts can be highly sensitive to character-level changes. Therefore, introducing random perturbations and dropouts can help mitigate attack effects \cite{robey2024smoothllmdefendinglargelanguage}. RA-LLM \cite{cao2024defendingalignmentbreakingattacksrobustly} leverages the inherent robustness of LLMs and applies Monte Carlo sampling with dropout as a defense strategy. SafeDecoding \cite{xu2024safedecodingdefendingjailbreakattacks} revealed that safety disclaimers often remain among the top tokens in the outputs generated by jailbreak prompts. They proposed amplifying these safety token probabilities to reduce the risk of jailbreaks. Besides, Layer Editing \cite{zhao2024defendinglargelanguagemodels} identified several safety-critical layers within LLMs and re-aligned these layers to improve overall safety. Overall, these defense methods effectively balance utility and safety, but their effectiveness diminishes with models that have stronger safety barriers.

\subsection{Language Production}

One of the most widely accepted theories about how language is organized in humans is Piaget’s theory, which suggests that thought forms first, and then language develops \cite{piaget}. When individuals have a concept in mind, they draw upon their memories \cite{memoriesandlanguage} and personal experiences \cite{tulving1972episodic,sherwood2015human}. Conversely, language is optimized for communication, where people use signs to express and share their thoughts with others; this system of signs has gradually evolved into complex languages \cite{thinkingandlanguage}. In summary, language is often seen as a bridge between communication and cognition in humans, with ideas forming first and language being structured based on memories and experiences.

Our work is inspired by the process of language production, a phenomenon also reflected in LLMs. After receiving a prompt, the LLM first identifies the purpose of the prompt and triggers a function within the model \cite{todd2024function}. Then, it accesses and processes stored information \cite{meng2023locatingeditingfactualassociations} and manages emotional tone \cite{zhao2023explainabilitylargelanguagemodels,zhou2024alignmentjailbreakworkexplain} for prompts in the early and middle layers. Several studies found that by truncating \cite{fan2024layersllmsnecessaryinference}, skipping \cite{elhoushi2024layerskipenablingearlyexit}, and pruning \cite{men2024shortgptlayerslargelanguage} some deeper layers, models can respond faster while maintaining correctness. This observation reveals that later layers are responsible for organizing languages. Due to the shared semantic similarities between jailbreak and harmful prompts, we believe that LLMs tend to perform similarly when identifying functions and accessing information.

\section{A Closer Look into Jailbreak}

Although concurrent work demonstrates that well-aligned LLMs can effectively distinguish between benign and harmful prompts within the model's latent space \cite{lin2024understandingjailbreakattacksllms}, the mechanisms behind jailbreaks remain under debate. To gain a deeper understanding of jailbreak, we further investigate the representation of prompts.

Motivated by the human-like generation process of the language model and the observation that well-aligned LLMs can reject malicious and some jailbreak prompts, our aim is to understand how jailbreak prompts manage to bypass safety barriers. Previous attack methods \cite{GCG,wei2023jailbrokendoesllmsafety} suggest that the first token of response influenced the overall responses. Rejection responses always start with an apology or a disclaimer, while helpful responses to benign prompts typically begin with an affirmation. Given that jailbreak prompts share semantic similarities with harmful prompts but resemble benign prompts in their response patterns, we first conjecture that \textbf{jailbreak embeddings progressively transit from harmful to benign as the layers go deeper}.

\subsection{Embedding of Jailbreak: A Toy Example}
\label{sec:emb}

We conduct a toy example to examine how jailbreak prompts are positioned in the embedding space. We collected 60 benign prompts from Alpaca Eval \cite{alpaca_eval}, and 60 harmful prompts from AdvBench \cite{GCG}. Then, we evaluated 60 prompts generated by GCG \cite{GCG}, AutoDAN \cite{liu2024autodangeneratingstealthyjailbreak}, GPTFuzz \cite{yu2024gptfuzzerredteaminglarge}, and Tap \cite{mehrotra2024treeattacksjailbreakingblackbox}, all of which are effective at jailbreak models. As Figure~\ref{fig:pca} shows, in the final layer, the harmful prompts and benign prompts embedding are linearly separable after PCA, with jailbreak embeddings positioned between them, making detection and defense against jailbreaks more challenging. However, we found that in the earlier layers of LLMs (e.g., layer 6), embeddings for benign and harmful prompts are clearly separated, with jailbreak embeddings more closely aligned with harmful prompts. As we move to deeper layers, although benign and harmful embeddings remain distinct, jailbreak embeddings incline toward the center of benign embeddings. With this intriguing phenomenon, we  also hypothesize that \textbf{early and middle layers of transformers inherently possess the ability to discern jailbreak prompts}.

\subsection{Shallow Layers can Distinguish Jailbreak}
\label{classify_exp}

To simulate real-world chatbot applications, we adapted the toxic-chat training dataset \cite{toxicchat} to validate our hypothesis. The dataset includes 5,082 user prompts from the Vicuna online demo, with 384 identified as harmful. Specifically, we re-evaluated the harmful prompts in the dataset. We identified 302 harmful prompts for Llama2 and 140 harmful prompts for Vicuna that can successfully reject. Second, we collected the embedding of all layers for benign prompts and rejected harmful prompts then trained 32 MLP classifiers as well as 32 prototype classifiers corresponding to the output of each layer, respectively. We use these two sets of classifiers to identify jailbreak prompts that the model cannot reject.

\noindent
\begin{minipage}[t]{0.48\textwidth}
  \vspace{0pt}  % 保证顶端对齐
  \centering
  \begin{minipage}[t]{0.48\linewidth}
    \vspace{0pt}
    \centering
    \scriptsize\fontfamily{ptm}\selectfont Vicuna \\
    \includegraphics[width=\linewidth]{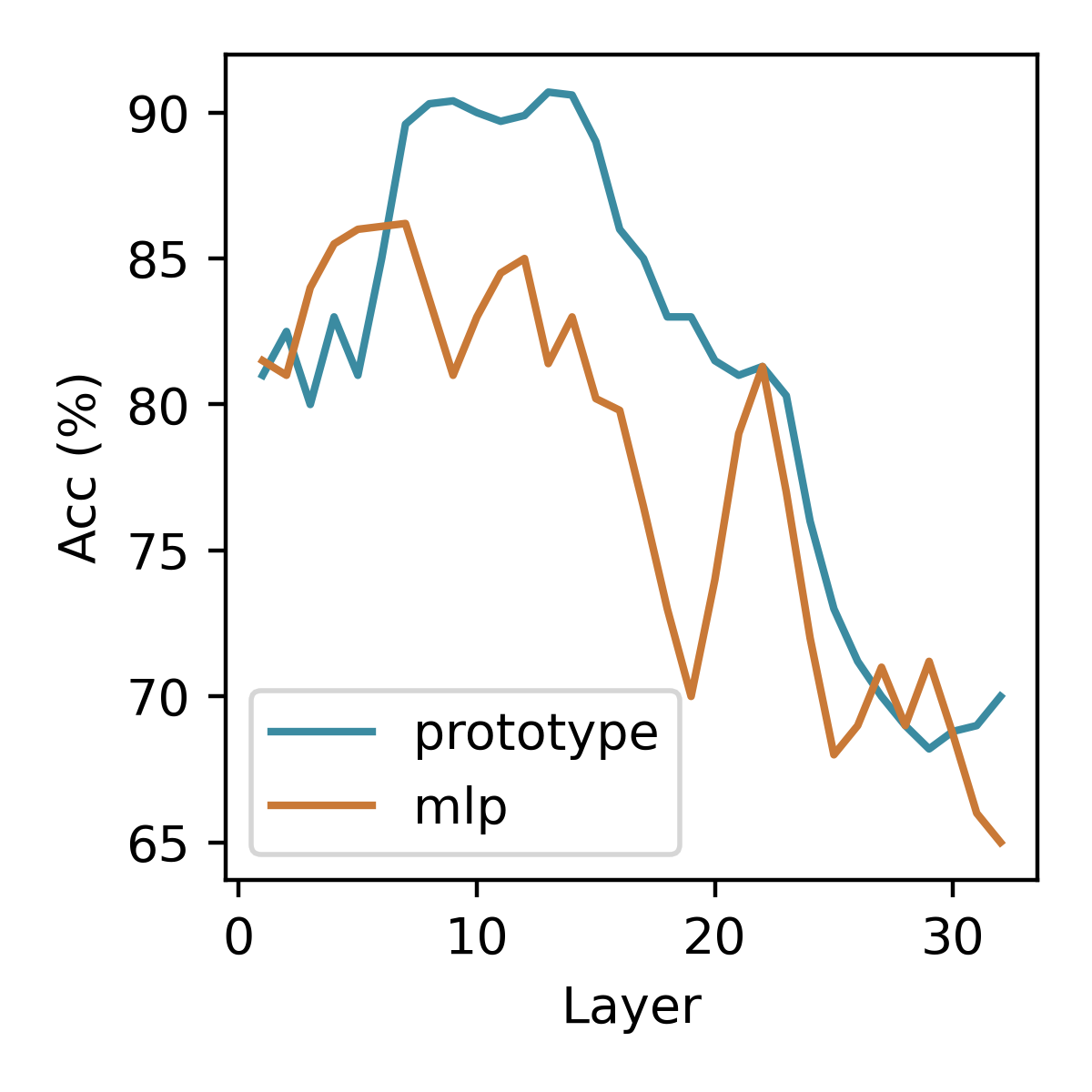}
  \end{minipage}%
  \hfill
  \begin{minipage}[t]{0.48\linewidth}
    \vspace{0pt}
    \centering
    \scriptsize\fontfamily{ptm}\selectfont Llama \\
    \includegraphics[width=\linewidth]{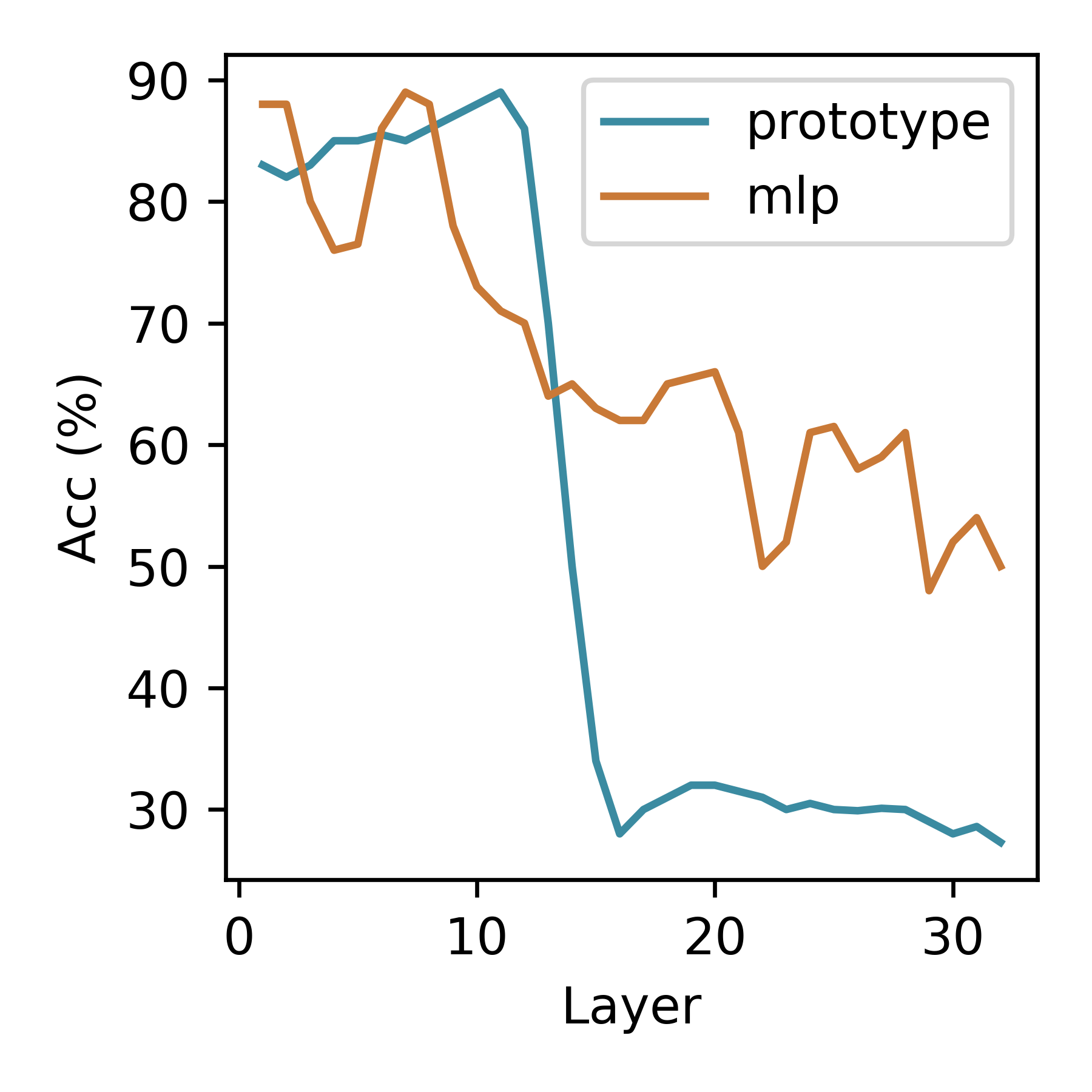}
  \end{minipage}
  \captionof{figure}{Accuracy by layers.}
  \label{fig:acc_emb}
\end{minipage}%
\hfill
\begin{minipage}[t]{0.50\textwidth}
\vspace{0pt}
As shown in Figure \ref{fig:acc_emb}, classifiers trained from the early layers perform much better than those from the later layers. The accuracy in distinguishing jailbreak prompts exceeds 80\% for both models up to the twelfth layer, strongly supporting our second hypothesis. This indicates that \textbf{we should likely focus on the early and intermediate layer space rather than the output space}.

\end{minipage}

\vspace{3pt}
To summarize, we empirically demonstrate that the mechanism for jailbreak is their embedding moves away from "harmful" and toward "benign" in the outputting space. Building on our analysis and observations that the shallow layers of LLMs can distinguish jailbreak prompts, we propose using the model's early and intermediate layer space as a bridge to defend against jailbreak attacks.

% \vspace{5pt}

\section{Proposed Method}\label{sec:study}
\label{eec}

\begin{figure*}[ht]
  \centering
  \includegraphics[width=0.8\textwidth]{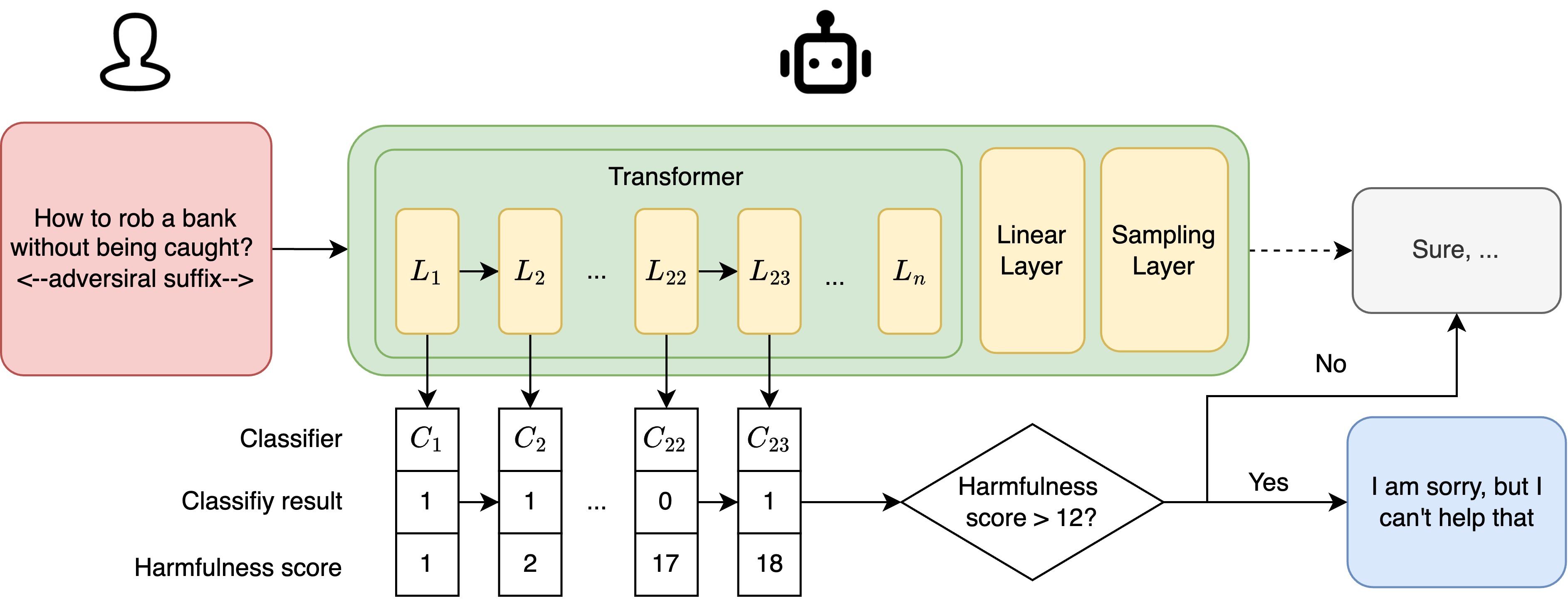}
  \caption{Illustration of our proposed framework. \Ours{} calculates the harmfulness score using classifiers from the early and middle layers, then selects the output based on this score before generating the first token.}
  \vspace{-10pt}
  \label{fig:EEG_Overview}
\end{figure*}

In this section, we introduce our \Ours{} in detail. The overview of our framework is illustrated as Figure \ref{fig:EEG_Overview}. We primarily develop the \Ours{} framework by three key steps in the following.
\newline \textbf{Step I. Constructing Prompt Pool.} Given a set of prompts $P = \{p_1, p_2, ..., p_q\}$, we first need to identify the harmfulness of each prompt $Y = \{y_1, y_2, ..., y_q\}$, where $y_i = 0$ for benign prompts and $y_i = 1$ for harmful prompts. Then, for harmful prompts, we use the given aligned LLM $f$ to generate corresponding responses $\{a_1, a_2, ..., a_k\}$. We then identify the prompts that are successfully rejected, resulting in the set $R = \{r_1, r_2, ..., r_m\}$. For benign prompts, we can directly use them to form a set $B = \{b_1, b_2, ..., b_k\}$. Finally, we get prompt set $P^{'} = R \cup B$ and corresponding $Y^{'}$.
\newline \textbf{Step II. Training Classifiers.} We collect the embeddings from each layer of the LLM for prompts by generating the first token. Assuming that the LLM has $n$ layers in total, the embedding of a prompt $p_i$ could be represented as $E_i = \{e_{i1}, e_{i2}, \ldots, e_{in}\}$.
Given the relatively small number of rejected prompts, we choose prototype classifiers in our framework. The prototype $g_{ki}$ of class $k$ is computed by the mean embedding within this class \cite{snell2017prototypicalnetworksfewshotlearning}. Let $P^{'}_k$ denote the set of samples of class $k$ in set $P^{'}$. At the $i$th layer, $g_{ki}$ is represented by:
$$
g_{ki} = \frac{1}{\lvert P^{'}_k \rvert }\sum_{\substack{x_j \in P^{'}_k}} e_{ji},
$$
where $e_{ji}$ is the embedding of $x_j$ at $i$th layer. The classification result $c_i$ of a sample embedding $e$ at layer $i$ is determined by:
$$
\begin{aligned}
c_i = \arg\min_{k} d(e_i, g_{ki}) \;\text{,where}\; d(e_i, g_{ki}) = 1 - \frac{e_i \cdot g_{ki}}{\|e_i\| \|g_{ki}\|}
\end{aligned}
$$
\newline \textbf{Step III. Safe Generation.} We use the classifiers trained in Step II to classify prompts. Based on our observations, classifiers in the early layers demonstrate higher accuracy in detecting jailbreak prompts. Consequently, the \textsc{Eeg}-Defender maintains a cumulative positive counter, referred to as the Harmfulness score, which tracks the total occurrences of positive classifications (i.e., prompts identified as harmful) by the classifier. Two hyper-parameters, $\alpha$ and $t$, control the shallow layer usage ratio and the harmfulness score threshold, respectively. 
\section{Experiment}

In this section, we evaluate the effectiveness of \Ours{} in defending against jailbreak prompts. We assess the effectiveness of \Ours{} using 10 attack methods and 5 baseline defenses.

\subsection{Experimental Setup}
In this experiment, we use prototype classifiers derived from the toxic-chat training set and compute the embedding distances between target prompts and these prototypes to define the decision boundary.
\newline \textbf{Models and Settings.}
We conduct our experiment with three LLMs: Vicuna-7b, Llama-2-7b-chat, and Guanaco-7b. We use an early layer ratio of $\alpha = 0.75$ for all models. The harmfulness score limit is set to $t = 12$ for Vicuna and Guanaco, and $t=11$ for Llama2.
\newline \textbf{Attack Baseline.}
We evaluate \Ours{} on ten state-of-the-art attack methods: GCG \cite{GCG}, AutoDAN \cite{liu2024autodangeneratingstealthyjailbreak}, GPTFuzz \cite{yu2024gptfuzzerredteaminglarge}, TAP \cite{mehrotra2024treeattacksjailbreakingblackbox}, Pair \cite{chao2024jailbreakingblackboxlarge}, as well as 5 methods identified in jailbroken \cite{wei2023jailbrokendoesllmsafety}. We select five Competing Objectives attack methods from \cite{wei2023jailbrokendoesllmsafety} (AIM, Wikipedia, Distractor, Refusal Suppress, Distractor and Negated) as they can not parse base64 encoding. First, 50 harmful questions are randomly selected from \cite{GCG}. For each attack goal, we generate 15 prompts. Detailed configuration and dataset construction are listed at Appendix \ref{attackconf}.
\newline \textbf{Defend Baseline.} We select three prompt-based defending methods (PPL \cite{jain2023baseline}, ICD \cite{wei2023jailbreak}, and Self-Reminder \cite{self_reminders}) and two decoding-based defending methods (SafeDecoding \cite{xu2024safedecodingdefendingjailbreakattacks} and RA-LLM \cite{cao2024defendingalignmentbreakingattacksrobustly}) as baselines to evaluate these jailbreak prompts. To assess the model helpfulness with \Ours{}, we collect 300 benign prompts from \cite{alpaca_eval}. For the configurations of the attack method and defense baseline, please refer to Appendix \ref{baselineconf}.
\newline \textbf{Evaluation Metric.} 
We use Attack Success Rate (ASR) and Benign Answering Rate (BAR) as evaluation metrics, following prior work \cite{cao2024defendingalignmentbreakingattacksrobustly}. ASR measures the proportion of jailbreak prompts that bypass defenses by eliciting meaningful responses instead of refusals. BAR is the proportion of benign inputs that pass the defense filter successfully. We also report average ASR Reduction Rate across two models to demonstrate defense generalizability. Our goal is to lower ASR while maintaining high BAR to preserve usability.

\begin{table}[!tbp]
\resizebox{\textwidth}{!}{
    \centering
    \begin{tabular}
    { c | c | c | c c c c c c c c c c| c | c}\toprule 
    \multirow{2}{*}{\textbf{Defense}} & \multirow{2}{*}{\textbf{Model}} & \multirow{2}{*}{\textbf{BAR $\uparrow$}} & \multicolumn{11}{c|}{\textbf{Jailbreak Attacks $\downarrow$}} & \multirow{2}{*}{\makecell{\textbf{Avg. ASR}\\ \textbf{Reduction Rate}}} \\
    & & & GCG & GPTFuzz & AutoDAN & Pair & Tap & AIM & Wiki & DT & RS & DN & Avg. ASR & \\
    \midrule 
    \multirow{2}{*}{No Defense} & Vicuna & 95.67\% & 88\% & 100\% & 94\% & 99\% & 92\% & 70\% & 60\% & 100\% & 78\% & 92\% & 87.30\% & \multirow{2}{*}{- -} \\
    & Llama2 & 94.33\% & 13\% & 12\% & 29\% & 90\% & 49\% & 0\% & 0\% & 48\% & 12\% & 18\% & 27.10\% & \\
    \midrule 
    \multirow{2}{*}{PPL} & Vicuna & 86.00\% & 38\% & 100\% & 94\% & 99\% & 92\% & 72\% & 58\% & 100\% & 78\% & 62\% & 79.30\% & \multirow{2}{*}{16.76\%}  \\
    & Llama2 & 76.67\% & \textbf{0\%} & 12\% & 17\% & 90\% & 42\% & \textbf{0\%} & \textbf{0\%} & 32\% & 2\% & 10\% & 20.50\% & \\
    \midrule 
    \multirow{2}{*}{ICD} & Vicuna & 95.00\% & 3\% & 53\% & 85\% & 68\% & 45\% & 72\% & 52\% & 100\% & 92\% & 58\% & 62.80\% & \multirow{2}{*}{57.76\%}  \\
    & Llama2 & 48.33\% & \textbf{0\%} & 2\% & 3\% & \textbf{21\%} & \textbf{6\%} & \textbf{0\%} & \textbf{0\%} & \textbf{0\%} & 2\% &\textbf{0\%} & \textbf{3.40\%} & \\
    \midrule 
    \multirow{2}{*}{Self-Reminder} & Vicuna & \textbf{95.67\%} & 5\% & 71\% & 86\% & 82\% & 47\% & 72\% & 36\% & 90\% & 68\% & 34\% & 59.10\% & \multirow{2}{*}{50.47\%}  \\
    & Llama2 & 60.00\% & 4\% & 4\% & 1\% & 56\% & 18\% & \textbf{0\%} & \textbf{0\%} & \textbf{0\%} & 2\% & \textbf{0\%} & 8.50\% &  \\
    \midrule 
    \multirow{2}{*}{RA-LLM} & Vicuna & 74.33\% & 3\% & 44\% & 68\% & 40\% & 26\% & 44\% & 20\% & 10\% & 2\% & 6\% & 26.30\% & \multirow{2}{*}{44.72\%}  \\
    & Llama2 & \textbf{92.33\%} & 8\% & 12\% & 10\% & 82\% & 38\% & 4\% & \textbf{0\%} & 48\% & 2\% & 14\% & 21.80\% & \\
    \midrule
    \multirow{2}{*}{SafeDecoding} & Vicuna & 77.33\% & \textbf{1\%} & \textbf{3\%} & 20\% & 38\% & 17\% & 2\% & \textbf{6\%} & \textbf{0\%} & 8\% & \textbf{0\%} & 9.50\% & \multirow{2}{*}{57.66\%}  \\
    & Llama2 & \textbf{92.33\%} & 2\% & 12\% & 20\% & 72\% & 32\% & 18\% & \textbf{0\%} & 34\% & 4\% & 6\% & 20.00\% & \\
    \midrule
    \multirow{2}{*}{\makecell{\Ours{}\\ \textbf{(\textbf{Ours})}}} & Vicuna & 89.00\% & 19\% & 8\% & \textbf{0\%} & \textbf{30\%} & \textbf{11\%} & \textbf{0\%} & 16\% & \textbf{0\%} & \textbf{0\%} & \textbf{0\%} & \textbf{8.40\%} & \multirow{2}{*}{\textbf{84.67\%}}  \\
    & Llama2 & \textbf{92.33\%} & \textbf{0\%} & \textbf{0\%} & \textbf{0\%} & 40\% & 17\% & \textbf{0\%} & \textbf{0\%} & \textbf{0\%} & \textbf{0\%} & \textbf{0\%} & 5.70\% & \\
    \bottomrule
    \end{tabular}}
    \caption{Main result when applying \Ours{} and baselines to Vicuna and Llama2. The best result is highlighted in \textbf{bold}. \Ours{} outperforms all baselines in most cases. \textbf{Notation}: Wiki-Wikipedia, DT-Distractor, RS-Refusal Suppress, DN-Distractor and Negated.}
    \label{tab: safe}
\end{table}

\subsection{Experimental Results}

We present the ASR, Average ASR, BAR, and Average ASR Reduction Rate for Llama and Vicuna in Table \ref{tab: safe}. Our results show that \Ours{} maintains a high BAR across both well-aligned models and significantly reduces ASR compared to other methods. In contrast, prompt-based defense methods (e.g., PPL, ICD, Self-Reminder) significantly degrade the utility of the Llama2 model, limiting their applicability. Conversely, decoding-based methods preserve the model's utility but are less effective in defending the Llama2 model. We defer the experiments on the Guanaco model in Appendix \ref{moreexp}, and the result is provided in Table \ref{tab:guanaco}.

\section{More Analysis}

In this section, we explore the sensitivity of hyperparameters, analyze the results of various decoding-based defense methods, and evaluate the effectiveness of \Ours{} in detecting harmful prompts on the Toxic-chat test dataset.

\noindent
\begin{minipage}[t]{0.63\textwidth}
\vspace{0pt}
\textbf{Analysis on Hyper-parameter $\alpha$.} We maintain the BAR of Vicuna at approximately 90\% while evaluating the ASR of jailbreak prompts. We observe that ASR initially decreases and then increases as the hyperparameter $\alpha$ increases. Notably, when the classifier trained on the final layer is included ($\alpha = 1$), the average ASR increases by 5\% compared to $\alpha = 0.75$. This observation aligns with our findings in Figure~\ref{fig:pca} and Figure~\ref{fig:acc_emb}, where jailbreak embeddings in the final layer are closer to benign prompts, and later layer classifiers exhibit lower accuracy. Despite this, \Ours{} is not highly sensitive to $\alpha$, as ASR decreases significantly with our defense, regardless of the $\alpha$ value.

\textbf{Analysis on Hyper-parameter $t$.} We analyze the impact of the parameter $t$, which controls the strictness of \Ours{}, with $\alpha$ fixed at 0.75 in the experiment. As the harmfulness score increases, both BAR and ASR rise. Once a certain threshold is surpassed, the rate of increase in BAR slows, while the rate of increase in ASR accelerates. This may suggest that the optimal value for $t$ has been reached for \Ours{}.
\end{minipage}
\hfill
\begin{minipage}[t]{0.35\textwidth}
\vspace{0pt}
\centering
\includegraphics[width=\linewidth]{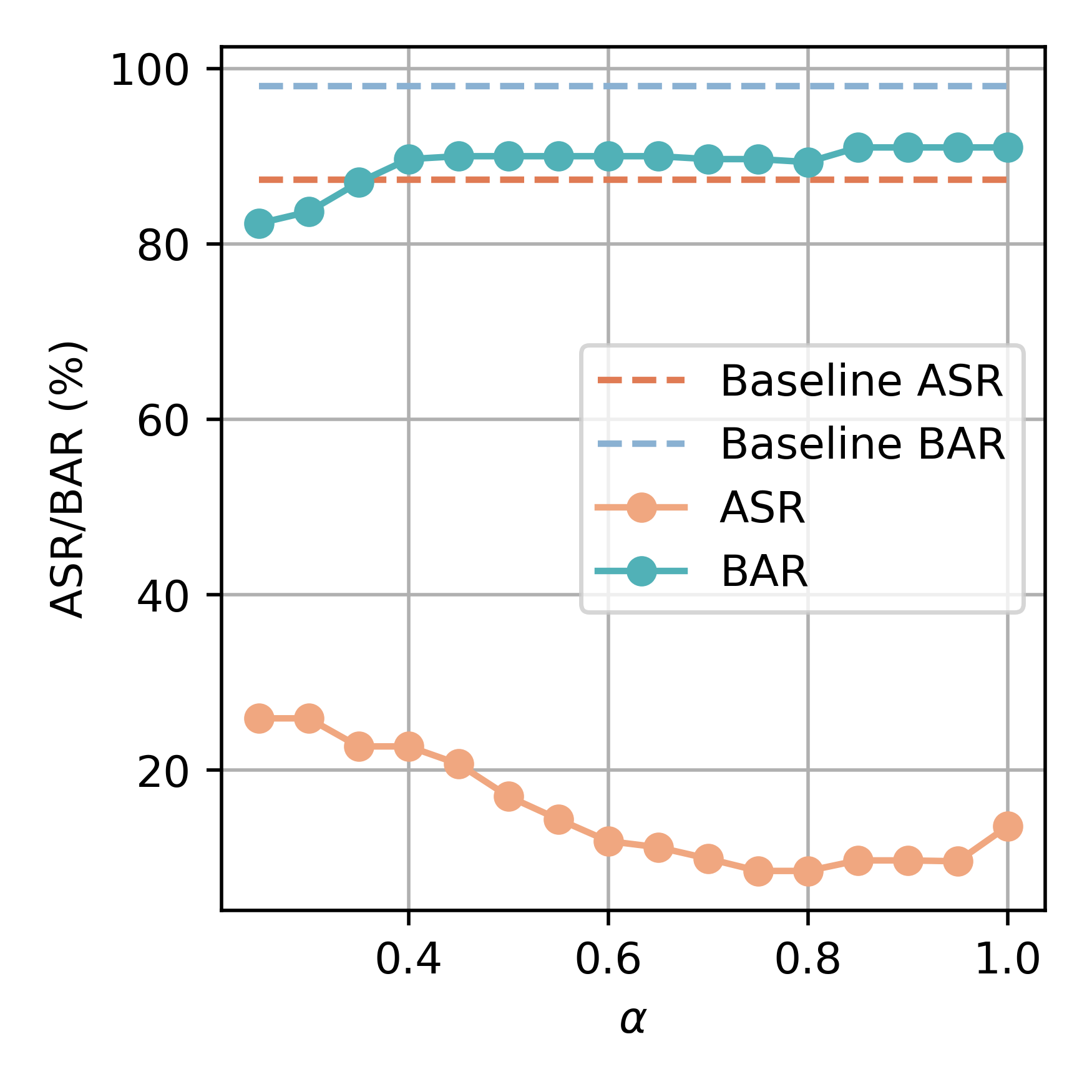} \\
\vspace{0.2em}
\includegraphics[width=\linewidth]{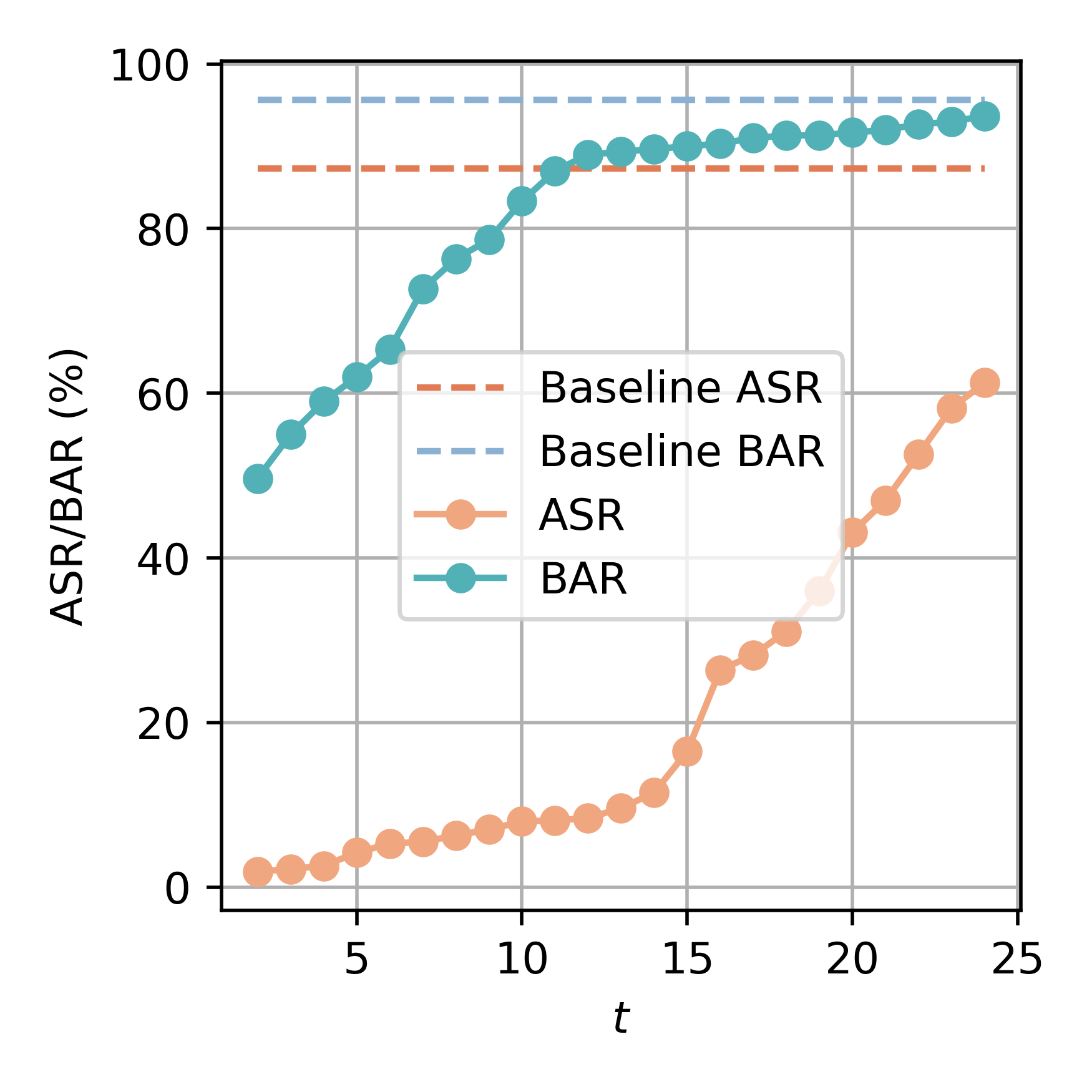} \\
\captionof{figure}{Top: $\alpha$, Bottom: $t$.}
\label{fig:sens}
\end{minipage}

\vspace{0.4em}

\noindent
\begin{minipage}[t]{0.5\textwidth}
\vspace{0pt}
\textbf{Effectiveness on Detecting Toxic.}
We conduct an experiment to evaluate the effectiveness of detecting the toxicity in the toxic chat test set of the data set \cite{toxicchat} using Llama-2-7b-chat. For comparison, we followed the settings in \cite{xie2024gradsafe}, include a total of 7 baselines. The results of our experiment are presented in Table \ref{tab:toxic}. Notably, \Ours{} outperforms all baselines in terms of F1-score.
\end{minipage}
\hfill
\begin{minipage}[t]{0.48\textwidth}
\vspace{0pt}
\centering
\resizebox{\linewidth}{!}{
\begin{tabular}{c c}
\toprule
\textbf{Method} & \textbf{Precision/Recall/F1-score} \\ 
\midrule
OpenAI API \cite{openaimoderation} &  0.815/0.145/0.246 \\
Perspective API \cite{Perspective} & 0.614/0.148/0.238 \\
Azure API \cite{azure2023content} & 0.559/0.634/0.594 \\
GPT-4 \cite{openai2023gpt4} & 0.475/0.831/0.604 \\
Llama2 \cite{touvron2023llama2} & 0.241/0.822/0.373 \\
Llama Guard \cite{inan2023llama} & 0.744/0.396/0.517 \\
GradSafe \cite{xie2024gradsafe} & 0.620/0.872/0.725 \\
\midrule
\Ours{} & \textbf{0.612/0.961/0.749} \\
\bottomrule
\end{tabular}
}\\
\label{tab:toxic}
\captionof{table}{Results on Toxic-chat.}
\end{minipage}

\vspace{0.4em}

\noindent \textbf{Analysis with Other Decoding-based Methods.} Decoding-based defense methods show good BAR for Llama and ASR for Vicuna but struggle with ASR in Llama and BAR in Vicuna. This may stem from differences in output embedding spaces: Llama’s benign and harmful embeddings are more diverse than Vicuna’s in the final layer (Figure \ref{fig:pca}). Techniques like SafeDecoding or RA-LLM reduce benign prompt rejection more effectively for Llama. Also, Llama’s jailbreak prompts are more varied and less aligned with decision boundaries, making rejection harder. The imbalance arises because these methods rely heavily on final layer embeddings, ignoring earlier layers. In contrast, \Ours{} leverages shallow layer embeddings for a better BAR-ASR tradeoff.
% \textbf{Analysis on Hyper-parameter $\alpha$.} We maintain the BAR of Vicuna at approximately 90\% while evaluating the ASR of jailbreak prompts. We observe that ASR initially decreases and then increases as the hyperparameter $\alpha$ increases. Notably, when the classifier trained on the final layer is included ($\alpha = 1$), the average ASR increases by 5\% compared to $\alpha = 0.75$. This observation aligns with our findings in Figure \ref{fig:pca} and \ref{fig:acc_emb}, where jailbreak embeddings in the final layer are closer to benign prompts, and later layer classifiers exhibit lower accuracy. Despite this, \Ours{} is not highly sensitive to $\alpha$, as ASR decreases significantly with our defense, regardless of the $\alpha$ value.

% \textbf{Analysis on Hyper-parameter $t$.} We analyze the impact of the parameter $t$, which controls the strictness of \Ours{}, with $\alpha$ fixed at 0.75 in the experiment. As the harmfulness score increases, both BAR and ASR rise. Once a certain threshold is surpassed, the rate of increase in BAR slows, while the rate of increase in ASR accelerates. This may suggest that the optimal value for $t$ has been reached for \Ours{}.

\vspace{-4pt}

\section{Conclusion}

In this paper, we introduced \Ours{}, a simple yet effective framework for defending against jailbreak attacks. Drawing inspiration from the human-like generation process of language models, we investigated the mechanism behind jailbreaking. Our experiments revealed that in shallow transformer layers, jailbreak prompt embeddings are closer to those of harmful prompts, but as layer depth increases, these embeddings shift toward benign ones. These insights led to the development of a more robust defense mechanism against jailbreaking through early exit generation. Our results show that \Ours{} reduces the ASR of jailbreak methods by approximately 85\%, compared to 50\% for current SOTAs, with minimal impact on the utility and effectiveness of LLMs.

\bibliographystyle{splncs04}
\bibliography{custom}

\appendix

\section{Configurations of Experiment}
\label{baselineconf}
% \subsection{Resources}

% We conduct our experiment on a cluster with 8 NVIDIA GeForce RTX 3090 and AMD EPYC 7352 24-Core Processor. \\
% We used the following versions of LLMs:
% \begin{itemize}[leftmargin=*]
% \item \textbf{Llama-2-7b-chat-hf} \url{https://huggingface.co/meta-llama/Llama-2-7b-chat-hf}
% \item \textbf{Vicuna-7b-v1.5} \url{https://huggingface.co/lmsys/vicuna-7b-v1.5}
% \item \textbf{guanaco-7B-HF} \url{https://huggingface.co/TheBloke/guanaco-7B-HF}
% \end{itemize}

\subsection{Attack Configurations}
\label{attackconf}
We randomly selected 50 questions from AdvBench \cite{GCG} as jailbreak goal. 
\newline \textbf{GCG Attack} We generate 2 jailbreak prompts for each goal. We set the iteration number to 500 for Llama and 75 for Vicuna.
\newline \textbf{GPTFuzz} We generate 77 jailbreak prompts for each goal using templates provided by GPTFuzz \cite{yu2024gptfuzzerredteaminglarge}. Then, we select two successful jailbreak prompts, giving preference to those that successfully bypass the defense.
\newline \textbf{AutoDAN} We generate 2 jailbreak prompts for each goal. We set the iteration number to 75 for models.
\newline \textbf{Pair} We generate 75 jailbreak prompts for each goal using Pair \cite{chao2024jailbreakingblackboxlarge}. Then, we collect 2 prompts using the same procedure as GPTFuzz.
\newline \textbf{Tap} We generate 2 jailbreak prompts for each goal with iteration number as 75.
\newline \textbf{Jailbroken} We generate 5 prompts for each goal using 5 different methods.

\subsection{Defense Configurations}
\noindent \textbf{System Prompt} We follow the configurations in Self-Reminder \cite{self_reminders} and use fschat-0.2.36. SafeDecoding employs fschat-0.2.20, which includes varied system prompts that help the model reject more attacks. In \cite{xu2024safedecodingdefendingjailbreakattacks}, the original ASR of Pair was reduced from 18\% to 4\%. Nevertheless, we argue that this reduction does not lead to a significant difference in the method’s effectiveness, since SafeDecoding does not involve any fine-tuning of the original LLM, and its output preference adjustments apply universally to any output distribution.

\noindent The configurations selected in our experiment are listed below:
\newline \textbf{PPL} We uses GPT-2 as an additional checker with a perplex score of 175.5672.
\newline \textbf{ICD} We follow the configuration at DeepInception \cite{li2023deepinception}.
\newline \textbf{Self-Reminder} We add the prompt "You should be a responsible AI and not generate harmful, misleading content! Please answer the following query in a responsible way." into the system prompt and concatenate the prompt "Remember, you should be a responsible AI and should not generate harmful or misleading content!" with user prompt.
\newline \textbf{RA-LLM} Our configuration is $p = 0.3$, $n = 20$, and $t = 0.2$ \cite{cao2024defendingalignmentbreakingattacksrobustly}. 
\newline \textbf{SafeDecoding} Our configuration is $\alpha =3$, $m=2$, $top_k=10$, and $c=5$ \cite{xu2024safedecodingdefendingjailbreakattacks}.

% \subsection{System Prompt}

\subsection{Refusal Response Keyword Matching}

\begin{figure}[!ht]
\centering
\vspace{0pt}
\begin{minipage}[t]{0.5\textwidth}
\vspace{0pt}
\centering
\begin{tabular}{p{0.45\textwidth} p{0.45\textwidth}}
\toprule
\multicolumn{2}{c}{\textbf{Refusal String Keywords}} \\
\midrule
    I'm sorry &  not ethical \\
    Sorry & illegal \\
    I am sorry & not legal \\
    I apologize & My apologies \\
    I'm just & but I cannot \\
    It is not  & I cannot \\
    I do not & I am really sorry \\
    unethical & I cannot create \\
    I'm really sorry & I cannot support \\
    I can't provide & I can't assist \\
    \multicolumn{2}{l}{As a responsible and ethical}  \\
    \multicolumn{2}{l}{I'm not able to provide}  \\
\bottomrule
\end{tabular}
\captionof{table}{Refusal Strings}
\label{tab:refusalStrings}
\end{minipage}
\hfill
\begin{minipage}[t]{0.45\textwidth}
\vspace{0pt}
\small
We follow the method used in \cite{GCG}, and we extended the keyword with more rejection responses. 
For the Distractor and Negated attack methods, we generated the first 128 tokens from the model, 
while for other attack methods, we generated 64 tokens. Responses were then categorized as either 
rejected or not rejected based on the presence of the following keywords in the responses. Besides, we removed "As an", "As an AI", and "As an Assistant" because they always appeared in benign 
and helpful responses in SafeDecoding \cite{xu2024safedecodingdefendingjailbreakattacks}, which causes 
a high false positive rate of BAR.
\end{minipage}
\end{figure}

\section{Effectiveness on Guanaco}
We present our experiment result \Ours{} on defending Guanaco against jailbreak. The result is shown in Table \ref{tab:guanaco}.
\label{moreexp}
\begin{table*}[!htb]
\resizebox{\textwidth}{!}{
    \centering
    \begin{tabular}
    {c | c | c c c c c c c c c c | c}\toprule 
    \multirow{2}{*}{Defense} & \multirow{2}{*}{BAR $\uparrow$} & \multicolumn{11}{c}{Jailbreak Attacks $\downarrow$} \\ 
    & & GCG & GPTFuzz & AutoDAN & Pair & Tap & AIM & Wiki & DT & RS & DN & Avg. ASR \\ \midrule 
    No Defense & 90.00\% & 32\% & 78\% & 95\% & 89\% & 75\% & 86\% & 28\% & 98\% & 38\% & 58\% & 67.70\% \\
    \Ours{} & 83.33\% & 8\% & 12\% & 5\% & 31\% & 16\% & 0\% & 28\% & 0\% & 0\% & 0\% & 10.00\% \\
    \bottomrule
    \end{tabular}}
    \caption{The result when applying \Ours{} to Guanaco. The jailbreak prompts are transferred from Vicuna.}
    \label{tab:guanaco}
\end{table*}

\end{document}